\documentclass[runningheads]{llncs}
\usepackage{graphicx}
\usepackage{comment}
\usepackage{amsmath,amssymb} 
\usepackage{color}

\usepackage{cite}
\usepackage{amsfonts}
\usepackage{algorithmic}
\usepackage{graphicx}
\usepackage{textcomp}
\usepackage{xcolor}
\usepackage{epsfig}
\usepackage{threeparttable}
\usepackage{amsmath}
\usepackage{dsfont} 
\usepackage{textcomp}
\usepackage{import}
\usepackage{graphics}
\usepackage{booktabs}
\usepackage{enumitem}
\usepackage{tabularx}
\usepackage{multirow}
\usepackage{amssymb}
\usepackage[utf8]{inputenc}

\begin{document}
\pagestyle{headings}
\mainmatter
\def\ECCVSubNumber{2607}  

\title{Location Sensitive Image Retrieval and Tagging}

\titlerunning{Location Sensitive Image Retrieval and Tagging}
%
\author{Raul Gomez\inst{1,2} \and
Jaume Gibert\inst{1} \and
Lluis Gomez\inst{2} \and
Dimosthenis Karatzas\inst{2}}
%
\authorrunning{R. Gomez et al.}
%
\institute{
Eurecat, Centre Tecnològic de Catalunya, Unitat de Tecnologies Audiovisuals\\
Computer Vision Center, Universitat Autònoma de Barcelona, Barcelona, Spain\\
\email{\{raul.gomez,jaume.gibert\}@eurecat.org \{lgomez,dimos\}@cvc.uab.es}} 
\maketitle
\begin{abstract}
People from different parts of the globe describe objects and concepts in distinct manners. Visual appearance can thus vary across different geographic locations, which makes location a relevant contextual information when analysing visual data. 
In this work, we address the task of image retrieval related to a given tag conditioned on a certain location on Earth.
We present {\normalfont LocSens}, a model that learns to rank triplets of images, tags and coordinates by plausibility, and two training strategies to balance the location influence in the final ranking. {\normalfont LocSens} learns to fuse textual and location information of multimodal queries to retrieve related images at different levels of location granularity, and successfully utilizes location information to improve image tagging.
\end{abstract}

\section{Introduction}
Image tagging is the task of assigning tags to images, referring to words that describe the image content or context. An image of a beach, for instance, could be tagged with the words \textit{beach} or \textit{sand}, but also with the words \textit{swim}, \textit{vacation} or \textit{Hawaii}, which do not refer to objects in the scene. On the other hand, image-by-text retrieval is the task of searching for images related to a given textual query. Similarly to the tagging task, the query words can refer to explicit scene content or to other image semantics. In this work we address the specific retrieval case when the query text is a single word (a tag).

\begin{figure}[t]
	\centering
  \includegraphics[width=1\linewidth]{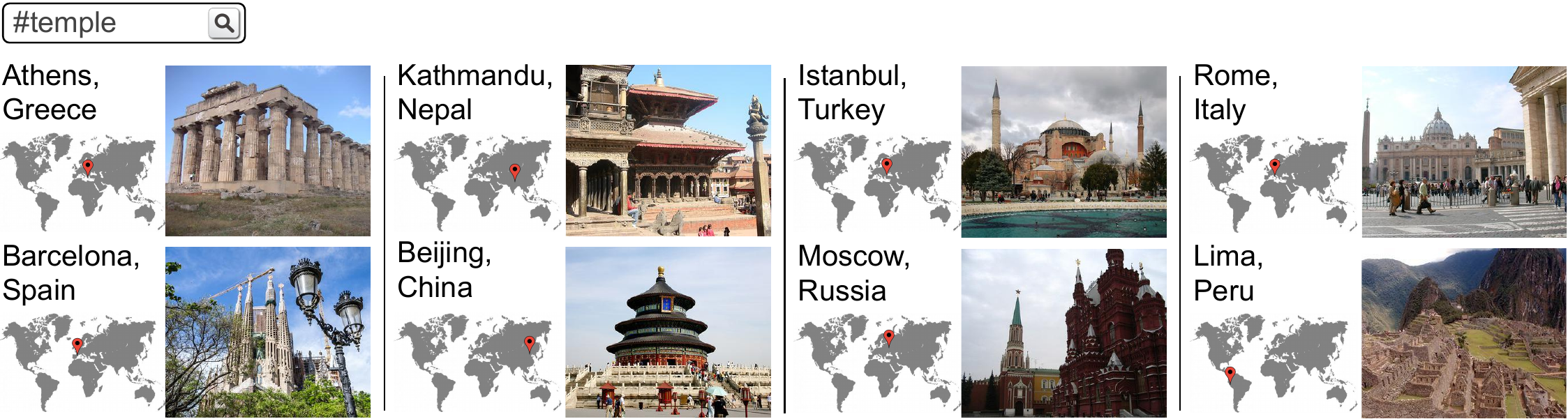}
  \vspace{-18pt}
   \caption{Top retrieved image by \textit{LocSens}, our location sensitive model, for the query hashtag ``\textit{temple}'' at different locations.}
   \label{fig:teaser}
   \vspace{-5.5mm}
\end{figure}

Besides text and images, location is a data modality widely present in contemporary data collections. 
Many cameras and mobile phones with built-in GPS systems store the location information in the corresponding \textit{Exif} metadata header when a picture is taken. Moreover, most of the web and social media platforms add this information to generated content or use it in their offered services.
In this work we leverage this third data modality: using location information can be useful in an image tagging task since location-related tagging can provide better contextual results.
For instance, an image of a skier in France could have the tags \textit{``ski, alps, les2alpes, neige''}, while an image of a skier in Canada could have the tags \textit{``ski, montremblant, canada, snow''}.
More importantly, location can also be very useful in an image retrieval setup where we want to find images related to a word in a specific location: the retrieved images related to the query tag \textit{temple} in Italy should be different from those in China. 
In this sense, it could be interesting to explore which kind of scenes people from different countries and cultures relate with certain \textit{broader} concepts. 
Location sensitive retrieval results produced by the proposed system are shown in Figure~\ref{fig:teaser}.

In this paper we propose a new architecture for modeling the joint distribution of images, hashtags, and geographic locations and demonstrate its ability to retrieve relevant images given a query composed by a hashtag and a location. In this task, which we call location sensitive tag-based image retrieval, a retrieved image is considered relevant if the query hashtag is within its ground-truth hashtags and the distance between its location and the query location is smaller than a given threshold. Notice that distinct from previous work on GPS-aware landmark recognition or GPS-Constrained database search \cite{Kennedy2007,Kennedy2008,OHare2005,Rattenbury2007} in the proposed task the locations of the test set images are not available at inference time, thus simple location filtering is not an option.

As an alternative to manually annotated datasets, Web and Social Media provide an interesting source of training data with all these sorts of modalities. The obvious benefits of this data are that it is free and virtually unlimited but also very diverse in terms of (weak) annotations, so we can find images with a huge diversity of hashtags and locations. 
Learning from this weakly supervised data, however, comes with challenges: missing labels, since an image is not necessarily annotated with hashtags referring to all of its contents, and noise, since users may tag images with hashtags that are not related to the image content.

A common approach to address these situations in both image by text retrieval and image tagging setups is to learn a joint embedding space for images and words \cite{Kiros2014,Wang2017a,Ren2016,Gomez2017}. In such a space, images are embedded near to the words with which they share semantics. Consequently, semantically \textit{similar} images are also embedded together. 
Usually, word embedding models, such as Word2Vec \cite{Mikolov2013} or GloVe \cite{Pennington} are employed to generate word representations, while a CNN is trained to embed images in the same space, learning optimal compact representations for them. Word models have an interesting and powerful feature: words with similar semantics have also similar representations and this is a feature that image tagging and retrieval models aim to incorporate, since learning a joint image and word embedding space with semantic structure provides a more flexible and less prone to drastic errors tagging or search engine.

Another approach to handle multiple modalities of data is by scoring tuples of multimodal samples aiming to get high scores on positive cases and low scores on negative ones \cite{Veit2018,Wang,Rohrbach,Jabri2016}.
This setup is convenient for learning from Web and Social Media data because, instead of strict similarities between modalities, the model learns more relaxed compatibility scores between them.
Our work fits under this paradigm. Specifically, we train a model that produces scores for image-hashtag-coordinates triplets, and we use these scores in a ranking loss in order to learn parameters that discriminate between observed and unobserved triplets. Such scores are used to tag and retrieve images in a location aware configuration providing good quality results under the large-scale YFCC100M dataset \cite{Thomee2015}. Our summarized contributions are:

\vspace{-1mm}
\begin{itemize}
    \setlength\itemsep{0em}
    \item We introduce the task of location sensitive tag-based image retrieval.
    \item We evaluate different baselines for learning image representations with hashtag supervision exploiting large-scale social media data that serve as initialization of the location sensitive model.
    \item We present the \textit{LocSens} model to score images, tags and location triplets (Figure~\ref{fig:model}), which allows to perform location sensitive image retrieval and outperforms location agnostic models in image tagging. 
    \item We introduce novel training strategies to improve the location sensitive retrieval performance of \textit{LocSens} and demonstrate that they are crucial in order to learn good representations of joint hashtag+location queries.
\end{itemize}
\vspace{-1mm}

\section{Related Work}
The computer vision and multimedia research communities have extensively explored the use of geotagged images for different applications~\cite{Luo2011,Min2016}, we discuss here the most related to our work. However, to the best of our knowledge, the task of location sensitive retrieval as defined before, has not yet been addressed.

{\bf Location-aware image search and tagging.}
O’Hare \textit{et al.} \cite{OHare2005} presented the need of conditioning image retrieval to location information, and targeted it by using location to filter out distant photos and then performing a visual search for ranking. Similar location-based filtering strategies have been also used for landmark identification~\cite{chen2011city} and to speed-up loop closure in visual SLAM~\cite{kumar2008experiments}. The obvious limitation of such systems compared to \textit{LocSens} is that they require geolocation annotations in the entire retrieval set. Kennedy \textit{et al.} \cite{Kennedy2007,Kennedy2008} and Rattenbury \textit{et al.} \cite{Rattenbury2007} used location-based clustering to get the most representative tags and images for each cluster, and presented limited image retrieval results for a subset of tags associated to a given location (landmark tags). They did not learn, however, location-dependent visual representations for tags as we do here, and their system is limited to the use of landmark tags as queries. 
On the other hand, Zhang \textit{et al.} \cite{Zhang2017a} proposed a location-aware method for image tagging and tag-based retrieval that first identifies points of interest, clustering images by their locations, and then represents the image-tag relations in each of the clusters with an individual image-tag matrix~\cite{wu2012tag}. Their study is limited to datasets on single city scale and small number of tags (1000). Their retrieval method is constrained to use location to improve results for tags with location semantics, and cannot retrieve location-dependent results (i.e. only the tag is used as query). Again, contrary to \textit{LocSens}, this method requires geolocation annotations over the entire retrieval set. Other existing location-aware tagging methods~\cite{moxley2008spirittagger,liu2014personalized} have also addressed constrained or small scale setups (e.g. a fixed number of cities) and small-size tag vocabularies, while in this paper we target a worldwide scale unconstrained scenario.

{\bf Location and Classification.}
The use of location information to improve image classification has also been previously explored, and has recently experienced a growing interest by the computer vision research community.  
Yuan \textit{et~al.} \cite{Yuan2008} combine GPS traces and hand-crafted visual features for events classification.
Tang \textit{et al.} \cite{Tang2015} propose different ways to get additional image context from coordinates, such as temperature or elevation, and test the usefulness of such information in image classification. 
Herranz \textit{et al.} \cite{Herranz2017,Xu2015} boost food dish classification using location information by jointly modeling dishes, restaurants and their menus and locations.
Chu \textit{et al.} \cite{Chu2019} compare different methods to fuse visual and location information for fine-grained image classification. 
Mac \textit{et al.} \cite{MacAodha2019} also work on fine-grained classification by modeling the spatio-temporal distribution of a set of object categories and using it as a prior in the classification process. Location-aware classification methods that model the prior distribution of locations and object classes can also be used for tagging, but they can not perform location sensitive tag-based retrieval because the prior for a given query (tag+location) would be constant for the whole retrieval set.

 {\bf Image geolocalization.}
Hays \textit{et al.} \cite{Hays} introduced the task of image geolocalization, i.e. assigning a location to an image, and used hand-crafted features to retrieve nearest neighbors in a reference database of geotagged images. 
Gallagher \textit{et al.} \cite{Gallagher2009} exploited user tags in addition to visual search to refine geolocalization.
Vo \textit{et al.} \cite{Vo2017} employed a similar setup but using a CNN to learn image representations from raw pixels.
Weyand \textit{et al.} \cite{Weyand} formulated geolocalization as a classification problem where the earth is subdivided into geographical cells, GPS coordinates are mapped to these regions, and a CNN is trained to predict them from images.
M\"uller-Budack \textit{et al.} \cite{Muller-Budack} enhanced the previous setup using earth partitions with different levels of granularity and incorporating explicit scene classification to the model.
Although these methods address a different task, they are related to \textit{LocSens} in that we also learn geolocation-dependent visual representations. Furthermore, inspired by \cite{Vo2017}, we evaluate our models’ performance at different levels of geolocation granularity.

{\bf Multimodal Learning.}
Multimodal joint image and text embeddings is a very active research area. DeViSE \cite{Frome} proposes a pipeline that, instead of learning to predict ImageNet classes, learns to infer the Word2Vec \cite{Mikolov2013} representations of their labels. 
This work inspired others that applied similar pipelines to learn from paired visual and textual data in a weakly-supervised manner~\cite{Gomez2017,DianeLarlus2017,Salvador}. More related to our work, Veit \textit{et al.} \cite{Veit2018} also exploit the YFCC100M dataset \cite{Thomee2015} to learn joint embeddings of images and hashtags for image tagging and retrieval. They work on user-specific modeling, learning embeddings conditioned to users to perform user-specific image tagging and tag-based retrieval. Apart from learning joint embeddings for images and text, other works have addressed tasks that need the joint interpretation of both modalities. Although some recent works have proposed more complex strategies to fuse different data modalities~\cite{Gao2018,Margffoy-Tuay2018,Vo2019,RajivJain2019,Xiaochang2019}, their results show that their performance improvement compared to a simple feature concatenation followed by a Multi Layer Perceptron is marginal.

\section{Methodology}
Given a large set of images, tags and geographical coordinates, our objective is to train a model to score triplets of image-hashtag-coordinates and rank them to perform two tasks: (1) image retrieval querying with a hashtag and a location, and (2) image tagging when both the image and the location are available.
We address the problem in two stages: first, we train a location-agnostic CNN to learn image representations using hashtags as weak supervision. We propose different training methodologies and evaluate their performance on image tagging and retrieval. 
These serve as benchmark and provide compact image representations to be later used within the location sensitive models. 
Second, using the learnt image and hashtags best performing representations and the locations, we train multimodal models to score triplets of these three modalities. We finally evaluate them on image  retrieval and tagging and analyze how these models benefit from the location information. 

\subsection{Learning with hashtag supervision}
\label{sec:models1}
Three procedures for training location-agnostic visual recognition models using hashtag supervision are considered: (1) multi-label classification, (2) softmax multi-class classification, and (3) hashtag embedding regression. 
In the following, let $\mathbb{H}$ be the set of $H$ considered hashtags. $\mathbf{I_x}$ will stand for a training image and $\mathbf{H_x} \subseteq \mathbb{H}$ for the set of its groundtruth hashtags. The image model $f(\cdot; \theta)$ used is a ResNet-50 \cite{He2016} with parameters $\theta$. 
The three approaches eventually produce a vector representation for an image $\mathbf{I_x}$, which we denote by $\mathbf{r_x}$. For a given hashtag $h^i \in \mathbb{H}$, its representation ---denoted $\mathbf{v_i}$--- is either learnt externally or jointly with those of the images.

\subsubsection{Multi-Label Classification (MLC).}
We set the problem in its most natural form: as a standard multi-label classification setup over $H$ classes corresponding to the hashtags in the vocabulary $\mathbb{H}$. The last ResNet-50 layer is replaced by a linear layer with $H$ outputs, and each one of the $H$ binary classification problems is addressed with a cross-entropy loss with sigmoid activation. 
Let $\mathbf{y_{x}} = (y_{x}^{1}, \dots, y_{x}^{H})$ be the multi-hot vector encoding the groundtruth hashtags of $\mathbf{I_x}$ and $\mathbf{f_x} = \sigma(f(\mathbf{I_{x}};\theta))$, where $\sigma$ is the element-wise sigmoid function. The loss for image $\mathbf{I_{x}}$ is written as:
\begin{equation}
\vspace{-4pt}
L = -\tfrac{1}{H} \sum_{h=1}^{H} [\, y_{x}^{h} \log f_{x}^{h} + (1-y_{x}^{h}) \log(1-f_{x}^{h}) \, ].
\vspace{-4pt}
\end{equation}

\subsubsection{Multi-Class Classification (MCC).}
Despite being counter-intuitive, several prior studies \cite{Mahajan,Veit2018} demonstrate the effectiveness of formulating multi-label problems with large numbers of classes as multi-class problems.
At training time a random target class from the groundtruth set $\mathbf{H_x}$ is selected, and softmax activation with a cross-entropy loss is used. This setup is commonly known as softmax classification. 

Let $h_{x}^{i}\in\mathbf{H_x}$ be a randomly selected class (hashtag) for $\mathbf{I_x}$.
Let also $f_{x}^{i}$ be the coordinate of $\mathbf{f_{x}}=f(\mathbf{I_{x}};\theta)$ corresponding to $h_{x}^{i}$. The loss for image $\mathbf{I_{x}}$ is set to be:
\begin{equation}
L = -\log \left ( \frac{e^{f_{x}^{i}}}{\sum_{j=1}^{H} e^{f_{x}^{j}}} \right ).
\end{equation}

In this setup we redefine ResNet-50 by adding a linear layer with $D$ outputs just before the last classification layer with $H$ outputs. This allows getting compact image $D$-dimensional representations $\mathbf{r_x}$ as their activations in such layer. 
Since we are in a multi-class setup where the groundtruth is a one-hot vector, we are also implicitly learning hashtag embeddings: the weights of the last classification layer with input $\mathbf{r_x}$ and output $\mathbf{f_x}$ is an $H \times D$ matrix whose rows can be understood as $D$-dimensional representations of the hashtags in $\mathbb{H}$. 
Consequently, this approach learns at once $D$-dimensional embeddings for both images and hashtags. 
In our experiments, the dimensionality is set to $D=300$ to match that of the word embeddings used in the next and last approach.
This procedure does not apply to MLC for which groundtruth is multi-hot encoded.

\subsubsection{Hashtag Embedding Regression (HER).}
We use pretrained GloVe \cite{Pennington} embeddings for hashtags, which are $D$-dimensional with $D=300$. For each image $\mathbf{I_{x}}$, we sum the GloVe embeddings of its groundtruth hashtags $\mathbf{H_{x}}$, which we denote as  $\mathbf{t_{x}}$. Then we replace the last layer of the ResNet-50 by a $D$-dimensional linear layer, and we learn the parameters of the image model by minimizing a cosine embedding loss. 
If, $\mathbf{f_{x}} = f(\mathbf{I_{x}};\theta)$ is the output of the vision model, the loss is defined by:
\begin{equation}
L = 1 - 
\left (
\frac 
{ \mathbf{t_{x}} \cdot  \mathbf{f_{x}} } 
{ \left \| \mathbf{t_{x}}\right \| \left \| \mathbf{f_{x}} \right \| }
\right )
.
\end{equation}

As already stated by \cite{Veit2018}, because of the nature of the GloVe semantic space, this methodology has the potential advantage of not penalizing predicting hashtags with close meanings to those in the groundtruth but that a user might not have used in the image description. Moreover, as shown in \cite{Frome} and due to the semantics structure of the embedding space, the resulting image model will be less prone to drastic errors.

\subsection{Location Sensitive Model (\textit{LocSens})}

We design a location sensitive model that learns to score triplets formed by an image, a hashtag and a location. We use a siamese-like architecture and a ranking loss to optimize the model to score positive triplets (existing in the training set) higher than negative triplets (which we create). Given an image $\mathbf{I_x}$, we get its embedding $\mathbf{r_x}$ computed by the image model, the embedding $\mathbf{v_{x_i}}$ of a random hashtag $h_{\mathbf{x}}^{i}$ from its groundtruth set $\mathbf{H_x}$ and its groundtruth latitude and longitude $\mathbf{g_x} = [\varphi_{\mathbf{x}}, \lambda_{\mathbf{x}}]$, which constitute a positive triplet. Both $\mathbf{r_x}$ and $\mathbf{v_{x_i}}$ are L2 normalized and latitude and longitude are both normalized to range in $[0,1]$. 
Note that $0$ and $1$ latitude fall on the poles while $0$ and $1$ represent the same longitude because of its circular nature and falls on the Pacific.

The three modalities are then mapped by linear layers with ReLU activations to $300$ dimensions each, and L2 normalized again. This normalization guarantees that the magnitudes of the representations of the different modalities are equal when processed by subsequent layers in the multimodal network. Then the three vectors are concatenated. Although sophisticated multimodal data fusion strategies have been proposed, simple feature concatenation has also been proven to be an effective technique \cite{Vo2019, Veit2018}. We opted for a simple concatenation as it streamlines the strategy.
The concatenated representations are then forwarded through $5$ linear layers with normalization and ReLU activations with $2048, 2048, 2048, 1024, 512$ neurons respectively. At the end, a linear layer with a single output calculates the score of the triplet. 
We have experimentally found that Batch Normalization \cite{Ioffe2015} hampers learning, producing highly irregular gradients. We conjecture that all GPU-allowable batch size is in fact a small batch size for the problem at hand, since the number of triplets is potentially massive and the batch statistics estimation will always be erratic across batches. Group normalization \cite{Wu2018} is used instead, which is independent of the batch size and permits learning of the models.
 
To create a negative triplet, we randomly replace the image or the tag of the positive triplet. The image is replaced by a random one not associated with the tag $h_{\mathbf{x}}^{i}$, and the tag by a random one not in $\mathbf{H_x}$. 
We have found that the performance in image retrieval is significantly better when all negative triplets are created replacing the image. This is because the frequency of tags is preserved in both the positive and negative triplets, while in the tagging configuration less common tags are more frequently seen in negative triplets.

\begin{figure}[h]
	\centering
	\vspace{-13pt}
  \includegraphics[width=0.5\linewidth]{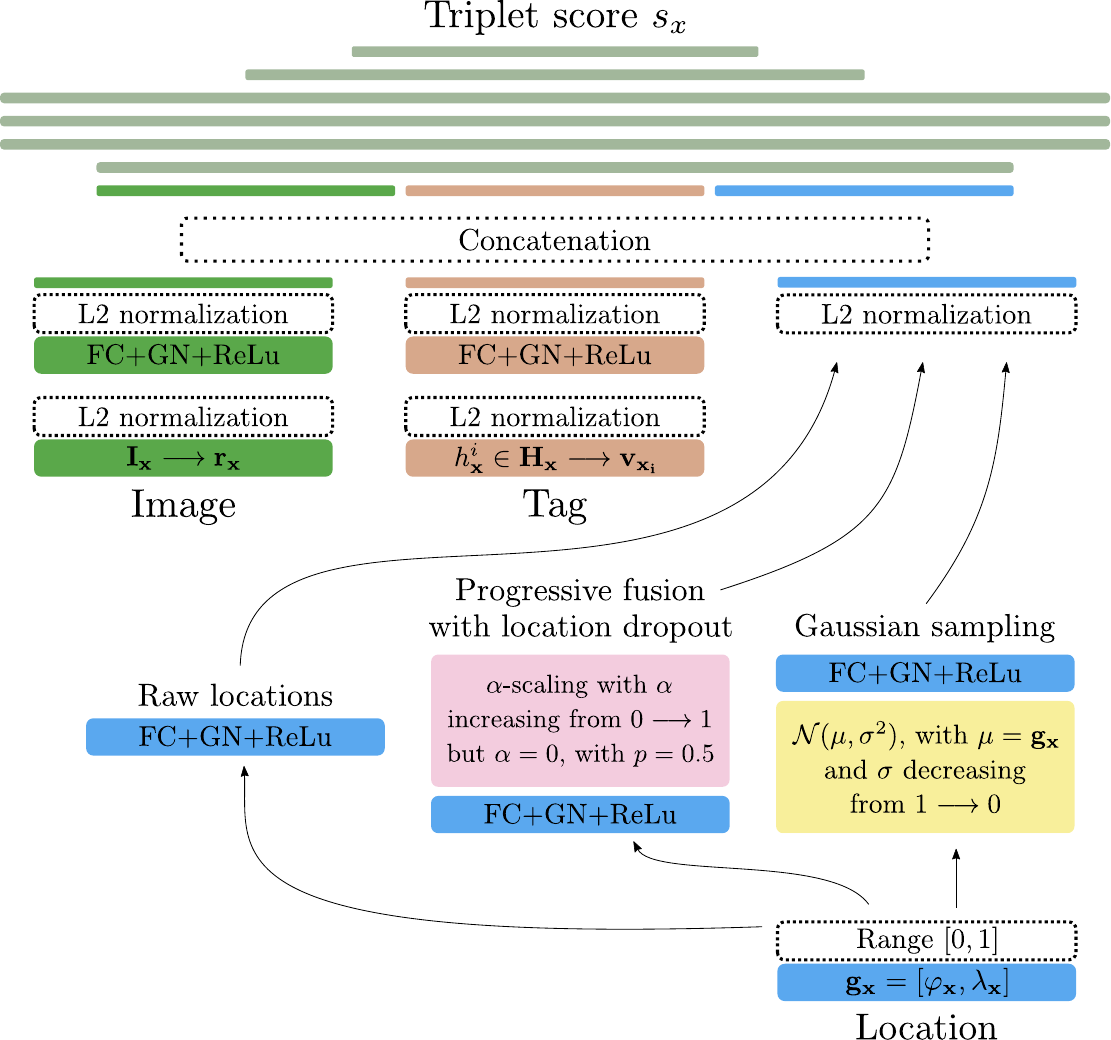}
  \vspace{-5pt}
   \caption{The proposed \textit{LocSens} multimodal scoring model trained by triplet ranking (bars after concatenation indicate fully connected + group normalization + ReLu activation layers). During training, location information is processed and inputted to the model with different strategies.
   }
   \label{fig:model}
   \vspace{-20pt}
\end{figure}

We train with a Margin Ranking loss, with a margin set empirically to $m = 0.1$, use $6$ negative triplets per positive triplet averaging the loss over them, and a batch size of $1024$. If $s_x$ is the score of the positive triplet and $s_n$ the score of the negative triplet, the loss is written as:
\begin{equation}
    L = max(0, s_n - s_x + m).
\end{equation}
Figure~\ref{fig:model} shows the model architecture and also the training strategies to balance location influence, which are explained next.

\subsubsection{Balancing Location Influence on Ranking.} 
\label{sec:balancing}
One important challenge in multimodal learning is balancing the influence of the different data modalities. We started by introducing the raw location values into the \textit{LocSens} model, but immediately observed that the learning tends to use the location information to discriminate between 
triplets much more than the other two modalities, forgetting previously learnt relations between images and tags. This effect is especially severe in the image retrieval scenario, where the model ends up retrieving images close to the query locations but less related to the query tag. This suggests that the location information needs to be gradually incorporated into the scoring model for location sensitive image retrieval. For that, we propose the following two strategies, also depicted in Figure~\ref{fig:model}.

\vspace{6pt}
\noindent\underline{Progressive Fusion with Location Dropout.} 
We first train a model with \textit{LocSens} architecture but silencing the location modality hence forcing it to learn to discriminate triplets without using location information. To do that, we multiply by $\alpha=0$ the location representation before its concatenation. 
Once the training has converged we start introducing locations progressively, by slowly increasing $\alpha$ until $\alpha=1$. This strategy avoids new gradients caused by locations to ruin the image-hashtags relations \textit{LocSens} has learned in the first training phase. 
In order to force the model to sustain the capability to discriminate between triplets without using location information we permanently zero the location representations with a $0.5$ probability. We call this \textit{location dropout} in a clear abuse of notation but because of its resemblance to zeroing random neurons in the well-known regularization strategy \cite{Srivastava2014}. For the sake of comparison, we report results for the \textit{LocSens} model with zeroed locations, which is in fact a location agnostic model.

\vspace{6pt}
\noindent\underline{Location Sampling.} Exact locations are particularly narrow with respect to global coordinates and such a fine-grained degree of granularity makes learning troublesome. We propose to progressively present locations from rough precision to more accurate values while training advances.   
For each triplet, we randomly sample the training location coordinates at each iteration from a $2D$ normal distribution with mean at the image real coordinates ($\mu = \mathbf{g_x}$) and with standard deviation $\sigma$ decreasing progressively. 
We constrain the sampling between $[0,\,1]$ by taking modulo $1$ on the sampled values.

We start training with $\sigma = 1$, which makes the training locations indeed random and so not informative at all. At this stage, the \textit{LocSens} model will learn to rank triplets without using the location information. Then, we progressively decrease $\sigma$, which makes the sampled coordinates be more accurate and useful for triplet discrimination. 
Note that $\sigma$ has a direct relation with geographical distance, so location data is introduced during the training to be first only useful to discriminate between very distant triplets, and progressively between more fine-grained distances. Therefore, this strategy allows training models sensitive to different location levels of detail.

\section{Experiments}
We conduct experiments on the YFCC100M dataset \cite{Thomee2015} which contains nearly 100 million photos from Flickr with associated hashtags and GPS coordinates among other metadata. We create the hashtag vocabulary following \cite{Veit2018}: we remove numerical hashtags and the 10 most frequent hashtags since they are not informative. The hashtag set $\mathbb{H}$ is defined as the set of the next 100,000 most frequent hashtags. Then we select photos with at least one hashtag from $\mathbb{H}$ from which we filter out photos with more than 15 hashtags. Finally, we remove photos without location information. This results in a dataset of 24.8M images, from which we separate a validation set of 250K and a test set of 500K. 
Images have an average of 4.25 hashtags.

\subsection{Image by Tag Retrieval}
\label{sec:exp1}
We first study hashtag based image retrieval, which is the ability of our models to retrieve relevant images given a hashtag query. We define the set of querying hashtags $\mathbb{H}^q$ as the hashtags in $\mathbb{H}$ appearing at least $10$ times in the testing set. The number of querying hashtags is $19,911$.
If $R_{h}^{k}$ is the set of top $k$ ranked images for the hashtag $h \in \mathbb{H}^q$ and $G_h$ is the set of images labeled with the hashtag $h$, we define precision$@k$ as:
\begin{equation}
P@k = \frac{1}{|\mathbb{H}^q|} \sum_{h\in \mathbb{H}^q } \frac{| R_{h}^{k} \cap G_h |} {k}.
\end{equation}
We evaluate precision$@10$, which measures the percentage of the $10$ highest scoring images that have the query hashtag in their groundtruth. Under these settings, precision$@k$ is upper-bounded by $100$. 
The precision$@10$ of the different location agnostic methods described in Section~\ref{sec:models1} is as follows: MLC:~1.01, MCC:~\textbf{14.07}, HER~(GloVe):~7.02. The Multi-Class Classification (MCC) model has the best performance in the hashtag based image retrieval task. 

\subsection{Location Sensitive Image by Tag Retrieval}
In this experiment we evaluate the ability of the models to retrieve relevant images given a query composed by a hashtag and a location (Figure~\ref{fig:teaser}). A retrieved image is considered relevant if the query hashtag is within its groundtruth hashtags and the distance between its location and the query location is smaller than a given threshold. Inspired by \cite{Vo2017}, we use different distance thresholds to evaluate the models' location precision at different levels of granularity. 
We define our query set of hashtag-location pairs by selecting the location and a random hashtag of $200,000$ images from the testing set. In this query set there will be repeated hashtags with different locations, and more frequent hashtags over all the dataset will also be more frequent in the query set (unlike in the location agnostic retrieval experiment of Section~\ref{sec:exp1}). This query set guarantees that the ability of the system to retrieve images related to the same hashtag but different locations is evaluated. To retrieve images for a given hashtag-location query with \textit{LocSens}, we compute triplet plausibility scores with all test images and rank them.

Table~\ref{tab:retrieval_sensitive_results} shows the performance of the different methods in location agnostic image retrieval and in different location sensitive levels of granularity. In location agnostic retrieval (first column) the geographic distance between the query and the results is not evaluated (infinite distance threshold). The evaluation in this scenario is the same as in Section~\ref{sec:exp1}, but the performances are higher because in this case the query sets contains more instances of the most frequent hashtags.
The upper bound ranks the retrieval images containing the query hashtag by proximity to the query location, showcasing the optimal performance of any method in this evaluation. In location sensitive evaluations the optimal performance is less than $100\%$ because we do not always have 10 or more relevant images in the test set.

\begin{table}[h]
\centering
\caption{\textbf{Location sensitive hashtag based image retrieval:} $P@10$. A retrieved image is considered correct if its groundtruth hashtags contain the queried hashtag and the distance between its location and the queried one is smaller than a given threshold}
\label{tab:retrieval_sensitive_results}
\resizebox{\textwidth}{!}{%
\begin{tabular}{p{10pt}lcccccc}
\toprule
\multicolumn{2}{c}{\textbf{}} & \multicolumn{6}{c}{\textbf{$P@10$}} \\
 \cmidrule{1-8}
& \textbf{Method} & \multicolumn{1}{c}{\textbf{\begin{tabular}[c]{@{}c@{}}Location \\ Agnostic\end{tabular}}} & \multicolumn{1}{c}{\textbf{\begin{tabular}[c]{@{}c@{}}Continent\\ (2500 km)\end{tabular}}} & \multicolumn{1}{c}{\textbf{\begin{tabular}[c]{@{}c@{}}Country\\ (750 km)\end{tabular}}} & \multicolumn{1}{c}{\textbf{\begin{tabular}[c]{@{}c@{}}Region\\ (200 km)\end{tabular}}} & \multicolumn{1}{c}{\textbf{\begin{tabular}[c]{@{}c@{}}City\\ (25 km)\end{tabular}}} & \multicolumn{1}{c}{\textbf{\begin{tabular}[c]{@{}c@{}}Street\\ (1 km)\end{tabular}}} \\
\toprule
& Upper Bound & 100 & 96.08 & 90.51 & 80.31 & 64.52 & 42.46 \\
\midrule
\parbox[t]{2mm}{\multirow{4}{*}{\rotatebox[origin=c]{90}{Img $+$ Tag}}} & MLC & 5.28 & 2.54 & 1.65 & 1.00 & 0.62 & 0.17 \\
& MCC & \textbf{42.18} & 29.23 & 24.2 & 18.34 & 13.25 & 4.66 \\
& HER (GloVe) & 37.36 & 25.03 & 20.27 & 15.51 & 11.23 & 3.65 \\
& LocSens - Zeroed locations & 40.05 & 28.32 & 24.34 & 18.44 & 12.79 & 3.74 \\
\midrule
\parbox[t]{2mm}{\multirow{7}{*}{\rotatebox[origin=c]{90}{Loc $+$ Img $+$ Tag}}} & LocSens - Raw locations & 32.74 & 28.42 & 25.52 & \textbf{21.83} & \textbf{15.53} & \textbf{4.83} \\
& LocSens - Dropout & 36.95 & 30.42 & 26.14 & 20.46 & 14.28 & 4.64 \\
& LocSens - Sampling $\sigma = 1$ & 40.60	& 28.40	& 23.84	& 18.16	& 13.04	& 4.13 \\
& LocSens - Sampling $\sigma = 0.1$ & 40.03 & 29.30	& 24.36	& 18.83	& 13.46	& 4.22\\
& LocSens - Sampling $\sigma = 0.05$ & 39.80 & 31.25& 25.76	& 19.58	& 13.78	& 4.30\\
& LocSens - Sampling $\sigma = 0.01$ & 37.05 & \textbf{31.27} & 26.65 & 20.14 & 14.15 & 4.44\\
& LocSens - Sampling $\sigma = 0$ & 35.95 & 30.61 & \textbf{27.00} & 21.39 & 14.75 &  \textbf{4.83}\\
\bottomrule
\end{tabular}}
\vspace{-10pt}
\end{table}

\begin{figure}[h]
	\centering
 \includegraphics[width=0.5\linewidth]{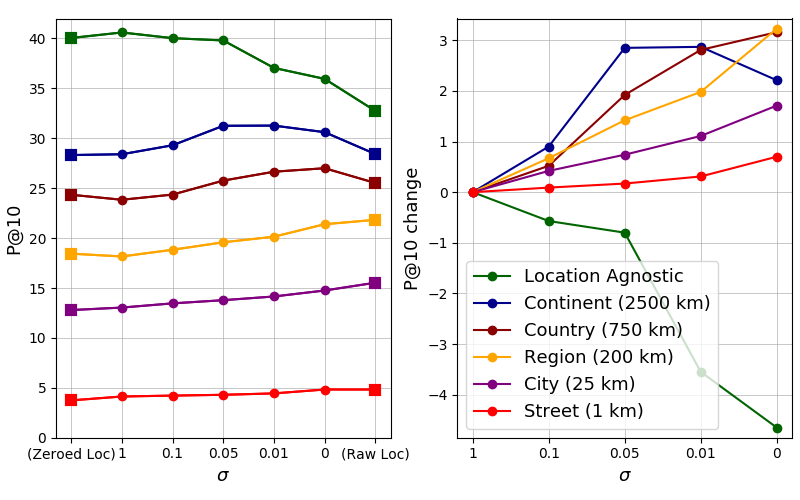}
 \vspace{-15pt}
  \caption{Left: $P@10$ of the location sampling strategy for different $\sigma$ and  models with zeroed and raw locations. Right: $P@10$ difference respect to $\sigma=1$.}
  \label{fig:variance_plot}
  \vspace{-15pt}
\end{figure}

Results show how the zeroed locations version of \textit{LocSens} gets comparable results as MCC. 
By using raw locations in the \textit{LocSens} model, we get the best results at fine level of location detail at the expense of a big drop in location agnostic retrieval. As introduced in Section~\ref{sec:balancing}, the reason is that it is relying heavily on locations to rank triplets decreasing its capability to predict relations between images and tags. As a result, it tends to retrieve images close to the query location, but less related to the query tag.
The proposed dropout training strategy reduces the deterioration in location agnostic retrieval performance at a cost of a small drop in the fine levels of granularity. Also, it outperforms the former models in the coarse continent and country levels, due to its better balancing between using the query tag and location to retrieve related images.
In its turn, the location sampling proposed approach with $\sigma=1$ gets similar results as \textit{LocSens} with zeroed locations because the locations are as irrelevant in both cases. When $\sigma$ is decreased, the model improves its location sensitive retrieval performance while maintaining a high location agnostic performance. This is achieved because informative locations are introduced to the model in a progressive way, from coarse to fine, and always maintaining triplets where the location is not informative, forcing the network to retain its capacity to rank triplets using only the image and the tag. 

Figure~\ref{fig:variance_plot} shows the absolute and relative performances at different levels of granularity while $\sigma$ is decreased.  At $\sigma = 0.05$, it can be seen that the location sensitive performances at all granularities have improved with a marginal drop on location agnostic performance. When $\sigma$ is further decreased, performances at finer locations keep increasing, while the location agnostic performance decreases. When $\sigma=0$, the training scenario is the same as in the raw locations one, but the training schedule allows this model to reduce the drop in location agnostic performance and at coarse levels of location granularity.

The location sampling technique provides \textit{LocSens} with a better balancing between retrieving images related to the query tag and their location. Furthermore, given that $\sigma$ has a direct geographical distance interpretation, it permits to tune the granularity to which we want our model to be sensitive.
Note that \textit{LocSens} enables to retrieve images related to a tag and near to a given location, which location agnostic models cannot do. The performance improvements in Table~\ref{tab:retrieval_sensitive_results} at the different levels of location granularity  are indeed significant since for many triplets the geographic location is not informative at all.

\begin{figure}[h]
	\centering
	  \vspace{-10pt}
  \includegraphics[width=1\linewidth]{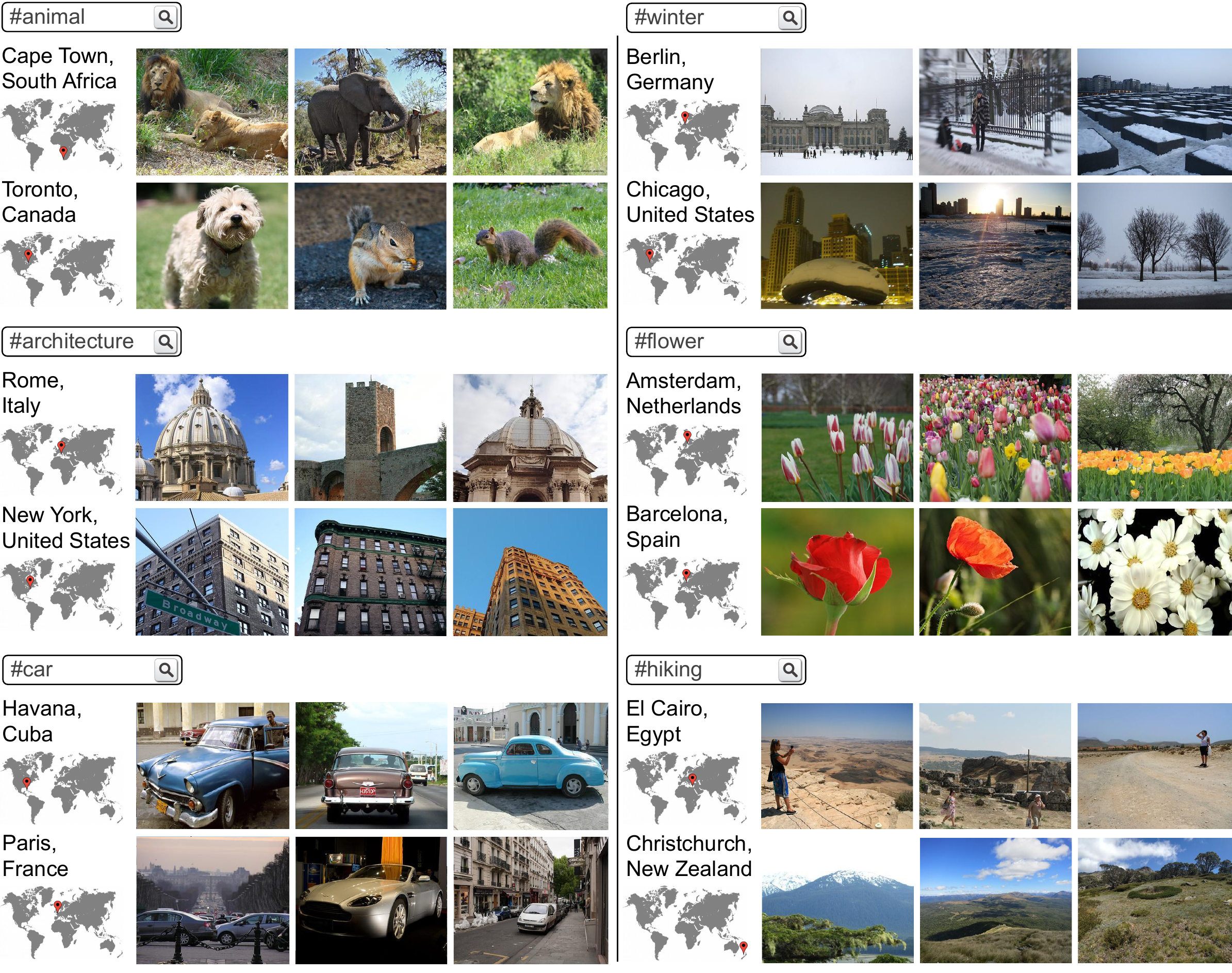}
  \vspace{-15pt}
   \caption{Query hashtags with different locations and top 3 retrieved images.}
   \label{fig:retrieval_results}
   \vspace{-15pt}
\end{figure}

Figures~\ref{fig:teaser} and \ref{fig:retrieval_results} show qualitative retrieval results of several hashtags at different locations. They demonstrate that the model successfully fuses textual and location information to retrieve images related to the joint interpretation of the two query modalities, being able to retrieve images related to the same concept across a wide range of locations with different geographical distances between them. \textit{LocSens} goes beyond retrieving the most common images from each geographical location, as it is demonstrated by the \textit{winter} results in Berlin or the \textit{car} results in Paris.


\subsection{Image Tagging}
In this section we evaluate the ability of the models to predict hashtags for images in terms of $A@k$ (accuracy at $k$). 
If $\mathbf{H_x}$ is the set of groundtruth hashtags of $\mathbf{I_x}$, $\mathbf{R_{x}^{k}}$ denotes the $k$ highest scoring hashtags for the image $\mathbf{I_x}$, and $N$ is the number of testing images, $A@k$ is defined as:
\begin{equation}
A@k = \frac{1}{N} \sum_{n=1}^{N} \mathds{1} \Big[ \mathbf{R_{n}^{k}} \cap \mathbf{H_n} \neq \emptyset \Big],
\end{equation}
where $\mathds{1}[\cdot]$ is the indicator function having the value of $1$ if the condition is fulfilled and $0$ otherwise.
We evaluate accuracy at $k=1$ and $k=10$, which measure how often the first ranked hashtag is in the groundtruth and how often at least one of the 10 highest ranked hashtags is in the groundtruth respectively. 

A desired feature of a tagging system is the ability to infer diverse and distinct tags \cite{Wu2017,Wu2018b}. In order to measure the variety of tags predicted by the models, we measure the percentage of all the test tags predicted at least once in the whole test set ($\%$\texttt{pred}) and the percentage of all the test tags correctly predicted at least once ($\%$\texttt{cpred}), considering the top 10 tags predicted for each image.

Table~\ref{tab:tagging_results} shows the performance of the different methods. 
Global Frequency ranks the tags according to the training dataset frequency. 
Among the location agnostic methods, MCC is the best one. This finding corroborates the experiments in \cite{Mahajan,Veit2018} verifying that this simple training strategy outperforms others when having a large number of classes.
To train the \textit{LocSens} model we used the image and tag representations inferred by the MCC model, since it is the one providing the best results. 

\begin{table}[h]
\centering
\vspace{-10pt}
\caption{\textbf{Image tagging:} $A@1$, $A@10$, $\%$\texttt{pred} and $\%$\texttt{cpred} of the frequency baseline, location agnostic prediction and the location sensitive model}
\label{tab:tagging_results}
\begin{tabular}{lrcrcrcrc}
\toprule
\textbf{Method} & \phantom{00} &\textbf{$A@1$} & \phantom{000} & \textbf{$A@10$} & \phantom{000} & $\%$\texttt{pred} & \phantom{00} & $\%$\texttt{cpred} \\
\midrule
Global Frequency    && 1.82              && 13.45             && 0.01            && 0.01  \\
\midrule
MLC                 && 8.86              && 30.59             && 8.04            && 4.5  \\
MCC                 && \textbf{20.32}    && \textbf{47.64}    && \textbf{29.11}  &&  \textbf{15.15} \\
HER (GloVe)         && 15.83             && 31.24             && 18.63           &&  8.74 \\
LocSens - Zeroed locations && 15.92      && 46.60             && 26.98           && 13.31 \\
\midrule
LocSens - Raw locations && \textbf{28.10} && \textbf{68.21}  &&  \textbf{44.00}  && \textbf{24.04} \\ 
\bottomrule
\end{tabular}
\vspace{-10pt}
\end{table}

To get the highest scoring tags for an image with location with \textit{LocSens}, we compute triplet plausibility scores with all the tags and rank them. \textit{LocSens - Zeroed locations} stands for a model where the location representations are zeroed, so it only learns to rank image and tag pairs. 
The aim of training this model is to check whether 
\textit{LocSens} additional capacity and training strategy are providing a boost on location agnostic tagging.
Results confirm they are not, since $A@10$ is comparable for both measures and $A@1$ drops significantly. This later deterioration is due to the softmax activations used in MCC, which foster highly frequent tags and penalize 
infrequent ones. Moreover, the training strategy of the \textit{LocSens} model does not penalize infrequent tags that much but suffers greatly from the missing labels problem.


\begin{figure}[h]
	\centering
  \includegraphics[width=1\linewidth]{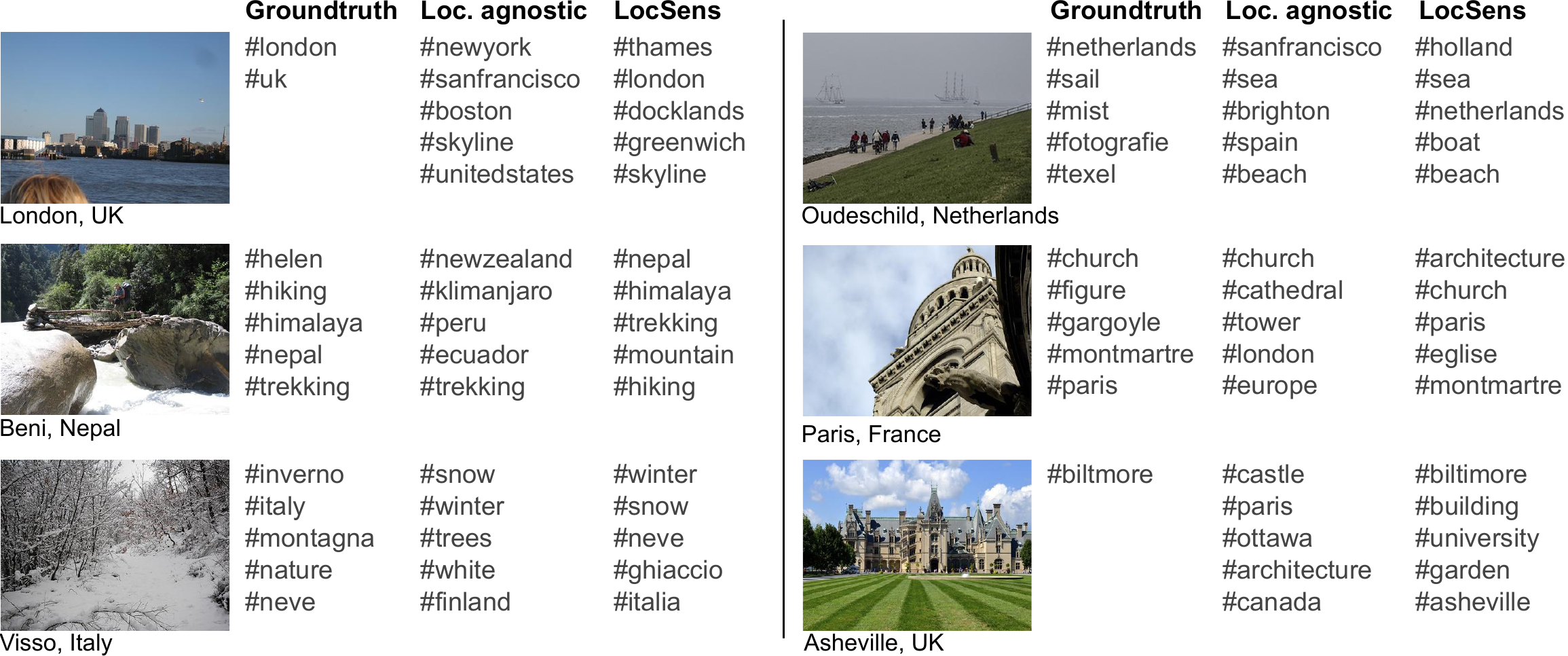}
  \vspace{-18pt}
   \caption{Images with their locations and groundtruth hashtags and the corresponding top 5 hashtags predicted by the location agnostic MCC model and \textit{LocSens}.}
   \label{fig:tagging_results}
\end{figure}

\begin{figure}[t]
	\centering
  \includegraphics[width=1\linewidth]{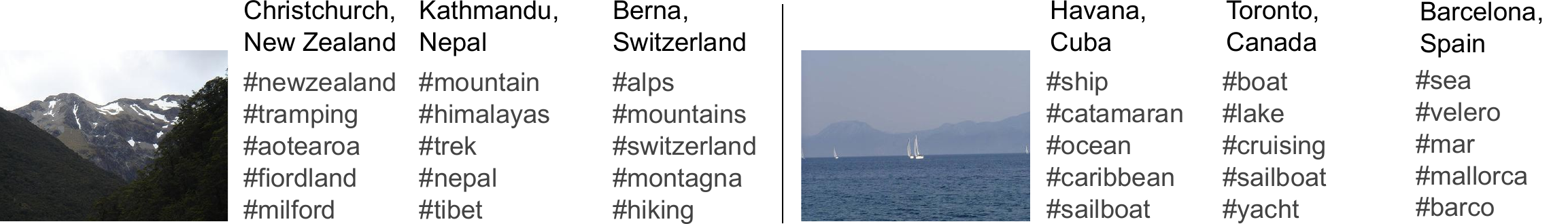}
  \vspace{-15pt}
   \caption{\textit{LocSens} top predicted hashtags for images with different faked locations.}
   \label{fig:tagging_results_multiple_locations}
   \vspace{-10pt}
\end{figure}

\textit{LocSens - Raw locations} stands for the model where the raw triplets locations are always inputted both at train and test time. It outperforms the location agnostic methods in accuracy, successfully using location information to improve the tagging results.
Moreover, it produces more diverse tags than location agnostic models, demonstrating that using location is effective for augmenting the hashtag prediction diversity.
Figure~\ref{fig:tagging_results} shows some tagging examples of a location agnostic model (MCC) compared to \textit{LocSens}, that demonstrate how the later successfully processes jointly visual and location information to assign tags referring to the concurrence of both data modalities.
As seen in the first example, besides assigning tags directly related to the given location (\textit{london}) and discarding tags related to locations far from the given one (\textit{newyork}), \textit{LocSens} predicts tags that need the joint interpretation of visual and location information (\textit{thames}). Figure~\ref{fig:tagging_results_multiple_locations} shows \textit{LocSens} tagging results on images with different faked locations, and demonstrates that \textit{LocSens} jointly interprets the image and the location to assign better contextualized tags, such as \textit{caribbean} if a sailing image is from Cuba, and \textit{lake} if it is from Toronto.

Note that \textit{LocSens} infers tags generally related to the image content while clearly conditioned by the image location, benefiting from the context given by both modalities. Tagging methods based solely on location, however, can be very precise predicting tags directly referring to a location, like placenames, but cannot predict tags related to the image semantics. We consider the later a requirement of an image tagging system, and we provide additional experimentation in the supplementary material of this work.



\section{Conclusions}
We have confirmed that a multiclass classification setup is the best method to learn image and tag representations when a large number of classes is available. Using them, we have trained \textit{LocSens} to rank image-tag-coordinates triplets by plausibility. 
We have shown how it is able to perform image by tag retrieval conditioned to a given location by learning location-dependent visual representations, and have demonstrated how it successfully utilizes location information for image tagging, providing better contextual results.
We have identified a problem in the multimodal setup, especially acute in the retrieval scenario: \textit{LocSens} heavily relies on location for triplet ranking and tends to return images close to the query location and less related to the query tag. 
To address this issue we have proposed two novel training strategies: progressive fusion with location dropout, which allows training with a better balance between the modalities influence on the ranking, and location sampling, which results in a better overall performance and enables to tune the model at different levels of distance granularity. 

%
%
\newpage
\bibliographystyle{splncs04}
\bibliography{MyCollection.bib}
\newpage

\pagebreak
\begin{center}
\textbf{\large Location Sensitive Image Retrieval and Tagging  Supplementary Material}
\end{center}
\setcounter{equation}{0}
\setcounter{figure}{0}
\setcounter{table}{0}
\setcounter{page}{1}

\section{LocSens vs Visual Agnostic Tagging}
\textit{LocSens} has the capability of jointly modeling visual and location information to assign better contextualized tags, and inferred tags are generally related to the image content while clearly conditioned by the image location, as shown in paper's Figures~3 and 4. 
However, image location by itself is a powerful information to infer tags, since the words with which users tag their images are highly dependent on location. In fact, in addition to tags related with image content, images are usually tagged with the name of the place where they were taken. Images with places names as tags are particularly common in the YFCC100M dataset used in this research, since most of the images are from photographer's travels which tend to tag their uploaded images with their travels destinations.
In this section we quantify how useful location information is if it is not jointly interpreted with visual information, and compare unimodal tagging performance with \textit{LocSens} performance. We then show how \textit{LocSens} goes beyond predicting places names, jointly interpreting visual and location information to assign better contextualized tags related to the image content.

\subsection{Location Based Baselines}
YFCC100M dataset provides also country, region and town names associated with each image, which have been specified by the user or inferred from the location. We computed the most frequent tags for each country and town in the training set. Then, we tagged each test image with the most common tags in its location to evaluate visual agnostic location based tagging baselines. Table~\ref{tab:tagging_results_sup} shows the performance of these baselines, the Multi-Class Classification location agnostic model and \textit{LocSens}. Location based baselines scores are high, and the \textit{Town Frequency} baseline outperforms the MCC (the best location agnostic baseline) in all metrics. It also outperforms \textit{LocSens} in A$@1$ and reaches a close score in A$@10$. However \textit{LocSens} A$@50$ score is superior by a large margin to unimodal models.
There are two reasons why location based baselines show high performances:

\vspace{6pt}
\noindent 1. Most of the YFCC100M images ($78\%$) are tagged with places names. Places names are actually among the most common tags in the dataset. For instance:
\vspace{4pt}

\tiny
Top global tags: london, unitedstates, england, nature, europe, japan, art, music, newyork, beach 

Top United States tags: unitedstates, newyork, sanfrancisco, nyc, washington, texas, florida, chicago, seattle

Top San Francisco tags: sanfrancisco, sf, unitedstates, francisco, san, iphone, protest, gay, mission 

\normalsize 

\vspace{4pt}

\noindent 2. In the A$@k$ metric it is enough to correctly infer one image tag to get the maximum score for that image. Therefore, since most of the images are tagged with places names, a tagging method solely based on location that does not predict tags related to the image content can get high scores. As an example, if an image is tagged with \textit{sydney}, \textit{beach}, \textit{sand} and \textit{dog}, a method predicting only \textit{sydney} from those tags would get the same A$@k$ score as a method predicting all of them. However, we use A$@k$ because is a standard performance metric for tagging and because it is also adequate to evaluate how \textit{LocSens} exploits location to outperform location agnostic models.
\vspace{6pt}

\textit{LocSens} outperforms the location baselines in A$@50$ by a big margin. One of the reasons is that \textit{LocSens} is also predicting correct tags for those images that do not have places names as tags.
 
\begin{table}[ht]
\centering
\caption{\textbf{Image tagging:} Accuracy$@1$, accuracy$@10$ and accuracy$@50$ of two visual agnostic hashtag prediction models, MCC and the location sensitive model.}
\label{tab:tagging_results_sup}
\begin{tabular}{lccc}
\toprule
\textbf{Method}             & \textbf{$A@1$}            & \textbf{$A@10$}       & \textbf{$A@50$}           \\
\midrule
Country Frequency   & 28.05  & 46.63  & 63.14    \\
Town Frequency   & \textbf{51.41}  & 65.49  &   71.05    \\
\midrule
MCC  & 20.32  & 47.64 & 68.05  \\
\midrule
LocSens - Raw locations   & 28.10 & \textbf{68.21} & \textbf{85.85} \\ 
\bottomrule
\end{tabular}
\end{table}

\subsection{Beyond Places Names}

Location based baselines achieve high A$@k$ scores by predicting places names as tags because most of the images are tagged with them. However, \textit{LocSens}, besides predicting tags related to image content and tags directly related to the given location, it predicts tags given the joint interpretation of visual and location information. To evaluate this behaviour, we omitted places names from groundtruth, frequency baselines and inferences and evaluated the methods. We construct the places list to omit by gathering all the continents, countries, regions and towns names in YFCC100M. Table~\ref{tab:tagging_results_noPlaces} shows the results. All performances are significantly worse, which is due to the less amount of groundtruth tags.
\textit{LocSens} performs much better than the best location agnostic model (MCC) even in this setup, where predicting places names tags is not evaluated. This proves that \textit{LocSens} goes beyond that, exploiting location information to jointly interpret visual and location information to predict better contextualized tags. In this case, \textit{LocSens} performs also much better than the location based baselines, since the reason of their high performance is their accuracy predicting places names, as explained in the former section.

\begin{table}[ht]
\centering
\caption{\textbf{Image tagging omitting places names:} Accuracy$@1$, accuracy$@10$ and accuracy$@50$ of two visual agnostic hashtag prediction models, MCC and the location sensitive model.}
\label{tab:tagging_results_noPlaces}
\begin{tabular}{lccc}
\toprule
\textbf{Method}             & \textbf{$A@1$}            & \textbf{$A@10$}       & \textbf{$A@50$}           \\
\midrule
Country Frequency   & 3.80 & 17.21  & 41.60    \\
Town Frequency   & 16.97 & 34.95  &   47.53    \\
\midrule
MCC  & 15.15  & 36.75 & 51.80  \\
\midrule
LocSens - Raw locations   & \textbf{17.34} & \textbf{44.45} & \textbf{61.10} \\ 
\bottomrule
\end{tabular}
\end{table}

\vspace{40pt}

\section{Results Analysis}

\subsection{Retrieval}

\subsubsection{Beyond retrieving common images at each location.}
\begin{figure}
	\centering
  \includegraphics[width=1\linewidth]{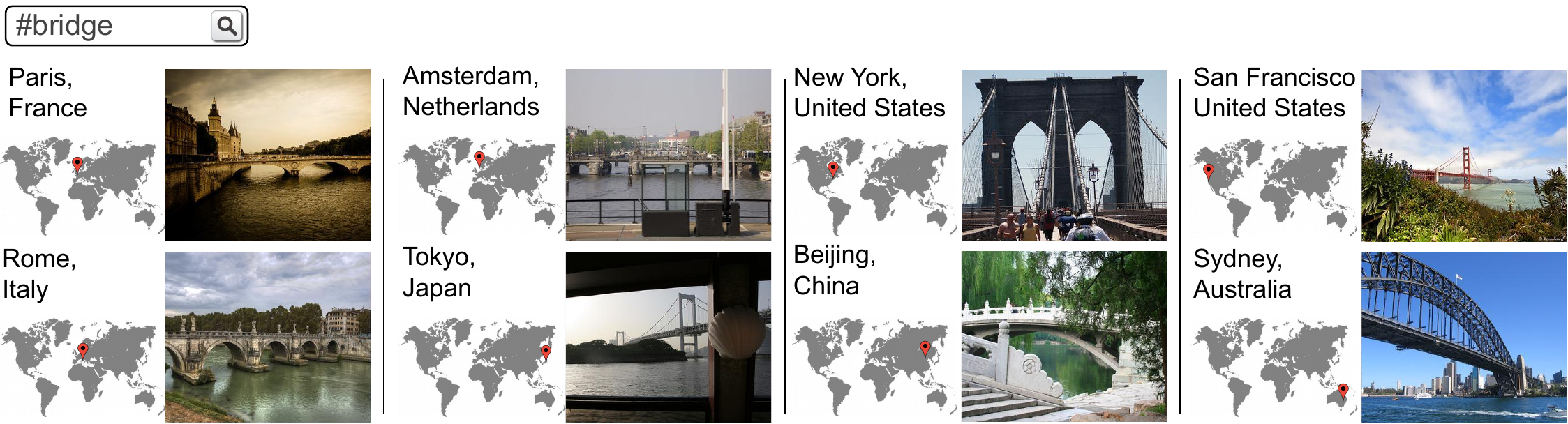}
   \caption{Top retrieved image by the location sensitive model for the query hashtag ``\textit{bridges}'' at different locations.}
   \label{fig:bridges}
\end{figure}

Paper's Figure~1 and Figure~\ref{fig:bridges} in this supplementary material show \textit{LocSens} retrieval results for the hashtag \textit{temple} and \textit{bridge} at different locations. They demonstrate how \textit{LocSens} is able to distinguish between images related to the same concept across a wide range of cities with different geographical distances between them. Note that, despite some specific bridges might have a huge amount of images tagged with \textit{bridge} in the dataset, as the San Francisco bridge or the Brooklyn bridge in New York, the system manages to retrieve images of other less represented bridges around the world. So, first and despite the bridges samples unbalance, it is learning to extract visual patterns that generalize to many different bridges around the world and, second, it is correctly balancing the tag query and location query influence in the final score.
Paper's Figure~5 shows \textit{LocSens} results for hashtags queries in different locations. The model is able to retrieve images related to a wide range of tags, from tags referring to objects, such as \textit{car}, to tags referring to more abstract concepts, such as \textit{hiking}, from the $100.000$ tags vocabulary. It goes beyond learning the most common images from each geographical location, as it is demonstrated by the \textit{hiking} results in El Cairo or the \textit{car} results in Paris, which are concepts that do not prevail in images in those locations, but the system is still able to accurately retrieve them.

\subsubsection{Challenging queries.}
\begin{figure}[h]
	\centering
  \includegraphics[width=1\linewidth]{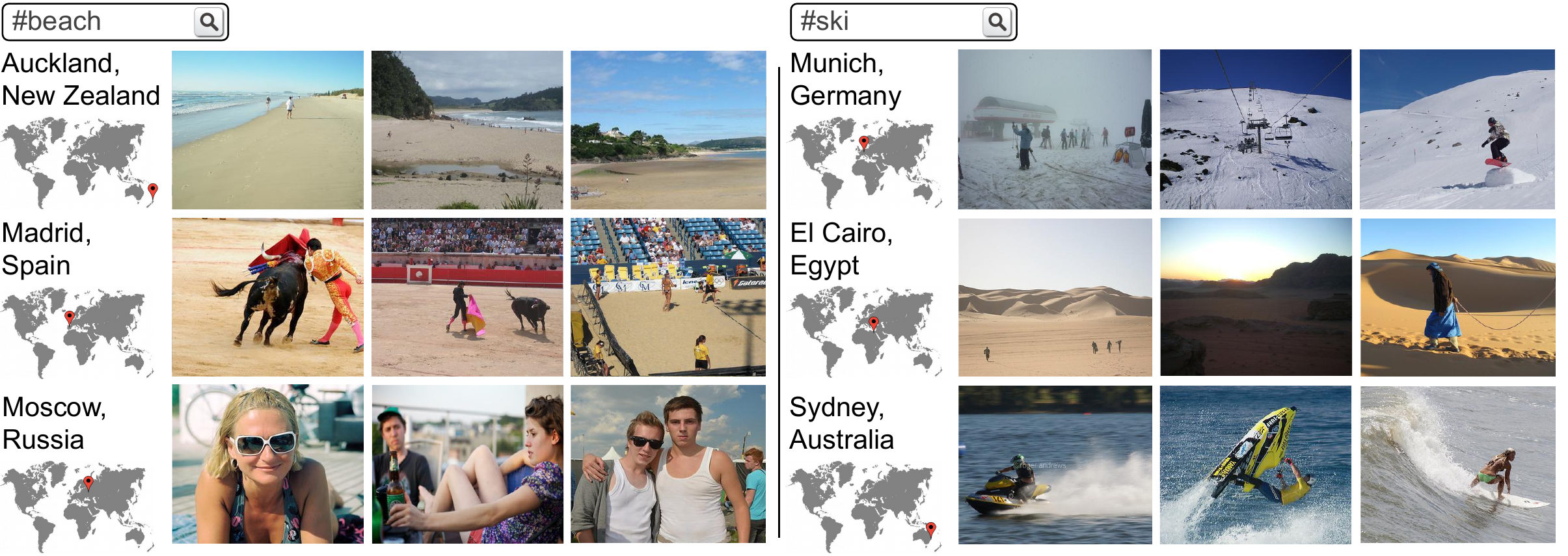}
   \caption{Query hashtags with different locations where some queries are incompatible because the hashtag refers to a concept which does not occur in the query location.}
   \label{fig:failures_retrieval}
\end{figure}

Figure~\ref{fig:failures_retrieval} shows \textit{LocSens} results for hashtag queries in different locations where some queries are incompatible because the hashtag refers to a concept which does not occur in the given location. When querying with the \textit{beach} hashtag in a coastal location such as Auckland, \textit{LocSens} retrieves images of close-by beaches. But when we query for \textit{beach} images from Madrid, which is far away from the coast, we get bullfighting and beach volley images, because the sand of both arenas makes them visually similar to beach images. If we try to retrieve beach images near Moscow, we get scenes of people sunbathing.
Similarly, if we query for \textit{ski} images in El Cairo and Sydney, we get images of the dessert and water sports respectively, which have visual similarities with ski images.

\subsubsection{P$@10$ depending on hashtag frequency.} 
\begin{figure}[h]
	\centering
  \includegraphics[width=0.5\linewidth]{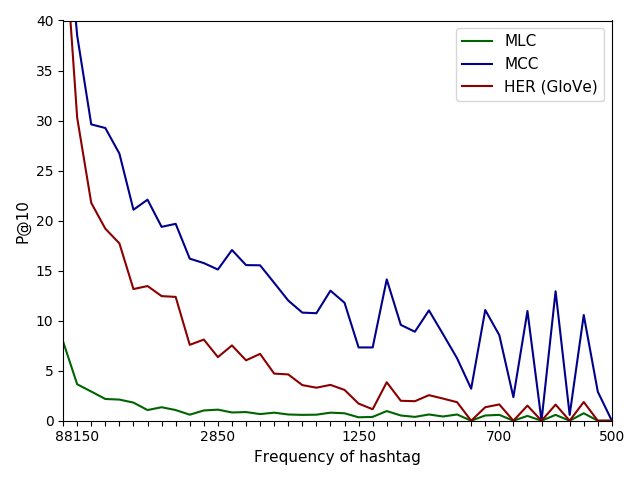}
   \caption{Image Retrieval P$@10$ per hashtag as a function of the number of hashtag appearances in the training set for the MLC, the MCC and the HER models.}
   \label{fig:P_by_freq}
\end{figure}

Figure~\ref{fig:P_by_freq} shows the P$@10$ score on location agnostic image retrieval for the MLC, the MCC and the HER training methods for query tags as a function of their number of appearances on the training set. It shows that all methods perform better for query hashtags that are more frequent in the training data, but MCC significantly outperforms the other methods also in less frequent hashtags.

\subsubsection{P$@10$ per continent at country granularity.} 
\begin{figure}[h]
	\centering
  \includegraphics[width=0.5\linewidth]{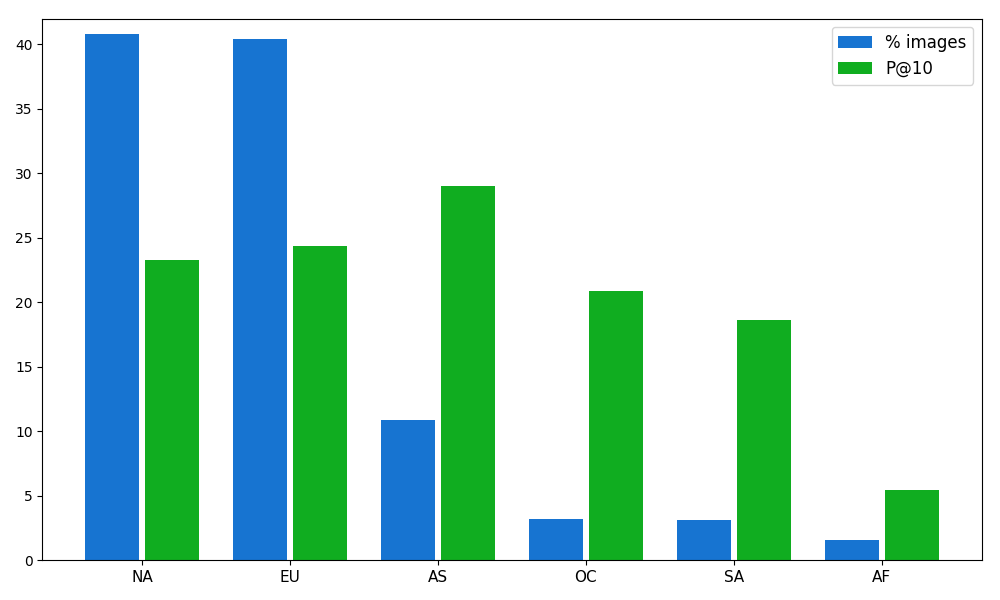}
   \caption{Number of training images per continent and Location Sensitive Image Retrieval P$@10$ at country granularity (750 km) per continent.}
   \label{fig:country_p_per_continent}
\end{figure}

Figure~\ref{fig:country_p_per_continent} shows the number of training images per continent, and the P$@10$ at country level (750 km) per continent of the \textit{LocSens} model performing better at it ($\sigma=0$). It shows how, with the exception of the Asia, the precision at country level is higher for continents with a bigger amount of training images.


\subsection{Tagging}

\begin{figure}[h]
	\centering
  \includegraphics[width=1\linewidth]{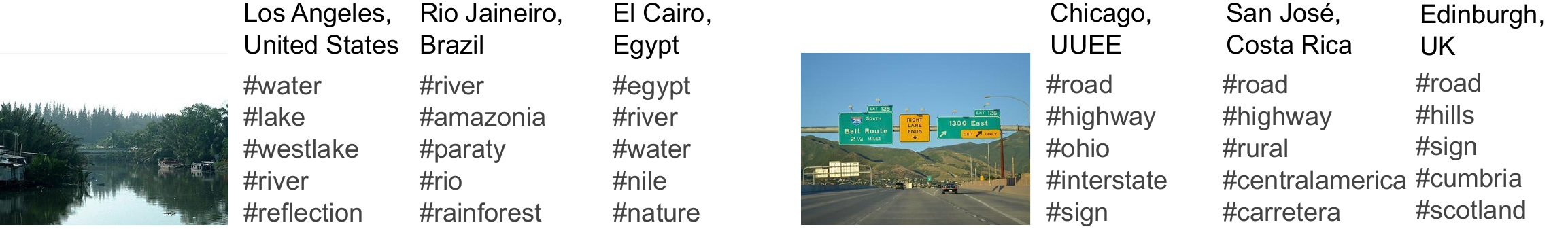}
   \caption{\textit{LocSens} top 5 predicted hashtags for images with 3 different faked locations.}
   \label{fig:tagging_results_multiple_locations_sup}
\end{figure}

Paper's Figure~4 and Figure~\ref{fig:tagging_results_multiple_locations_sup} of this supplementary material show \textit{LocSens} tagging results for images with different faked locations. They demonstrate that \textit{LocSens} is able to exploit locations to assign better contextualized tags, jointly interpreting both query visual and location modalities. For instance, it assigns to the river image \textit{lake} and \textit{westlake} if it is from Los Angeles, since Westlake is the nearest important water geographic accident, while if the image is from Rio de Janeiro it tags it with \textit{amazonia} and \textit{rainforest}, and with \textit{nile} if it is from El Cairo. 
In the example of an image of a road, it predicts as one of the most probable tags \textit{carretera} (which means \textit{road} in spanish) if the image is from Costa Rica, while it predicts \textit{hills}, \textit{Cumbria} and \textit{Scotland} if the image is from Edinburgh, referring to the geography and the regions names around. If the image is from Chicago, it predicts \textit{interstate}, since the road in it may be from the United States interstate highway system.
These examples prove the joint interpretation of the visual and the location modalities to infer the most probable tags, since predicted tags are generally related to the image content while clearly conditioned by the image location.

\section{Location Relevance in Image Retrieval}
The reason why the P@10 score difference between MCC and \textit{LocSens} on location sensitive image retrieval (shown in Table 1) is small is because the location information is not useful for many queries in our set because of their hashtags. There are several reasons for which a query hashtag can make the query location conditioning useless:

\begin{itemize}
     
\item\textbf{Hashtags carrying explicit location information.} Query hashtags that carry explicit location information are numerous in our query set, given it contains many travel pictures (i.e New York, Himalaya, Amazonas). See most frequent tags in the first section of this supplementary material.
\vspace{3pt}
\item\textbf{Hashtags carrying implicit location information.} Query hashtags that do not refer to specific locations, but carry implicit information of it. For instance, the language of the hashtag can indicate its location. Also hashtags referring to local celebrations, local dishes, etc.
\vspace{3pt}
\item\textbf{Hashtags with a visual appearance invariant to locastion}. Query hashtags that have the same visual appearance worldwide (such as “cat” or “tomato”), for which location-specific image features cannot be learnt. 

\end{itemize}

Therefore, the performance improvements of LocSens compared to MLC reported on Table 1 are small because location is irrelevant in many queries of this particular dataset, so LocSens is only able to outperform MLC in a small percentage of them. Besides, although MCC and \textit{LocSens} P@10 might be close, are qualitatively different an they do no retrieve the same images: As an example, LocSens - Raw locations retrieves images that are always near to the query location, but gets worse results than MCC in continent and country granularities because their relation with the query tag is weaker.
In this work we have focused on learning from large scale Social Media data. Further experimentation under more controlled scenarios where the location information is meaningful in all cases is another interesting research setup to evaluate the same tasks.

\section{Implementation Details}

\subsection{MLC}
The training of the MLC model was very unstable because of the class imbalance. We did try different class-balancing techniques without consistent improvements, and concluded that it is not an adequate training setup for our problem. We also tried different methods to evaluate both image tagging and retrieval using the MLC method, such as directly ranking the tags or the images with the scores, or learning embeddings with an intermediate 300-d layer as we do with MCC, but all experiments led to poor results.

\subsection{LocSens} 
\textit{LocSens} is trained with precomputed images and tag embeddings to reduce the computational load. Also, given LocSens has as inputs images but also tags embeddings learned by MCC, an architecture jointly optimizable would not be straight forward. 

\textit{LocSens} maps the image, tag, and location modalities to 300-d representations and then concatenates them. We experimented merging 2-d locations with the other modalities but couldn’t optimize \textit{LocSens} properly. We tried different strategies such as initializing LocSens parameters to attend location values, but mapping the three modalities to the same dimensionality before their concatenation yielded the best results. This is probably because it allows the model to better balance the different modalities. 

To train \textit{LocSens} with the location sampling technique we start always form $\sigma = 1$ and slowly decrease it to get models sensitive to different location granularities, evaluated in Table 1..

\section{Future Work}
The presented work can give rise to further research on how to exploit location information in image retrieval and tagging tasks, and also on how to learn image representations with tags supervision from large scale weakly annotated data.  We spot three different experimentation lines to continue with this research work:

\begin{itemize}
    
\item\textbf{Learning with tags supervision.} Our research on learning image representations with hashtags supervision concludes that a Multi-Class setup with Softmax activations and a Cross-Entropy loss outperforms the other baselines by a big margin. A research line to uncover the reason for this superior performance and to find under which conditions this method outperforms other standard learning setups, such as using a Multi-Label setup with Sigmoid activations, would be very interesting for the community.
\vspace{3pt}
\item\textbf{More efficient architectures.} The current efficiency of the method is a drawback, since for instance to find the top tags for an image and location query, we have to compute the score of the query with all the hashtags in the vocabulary. An interesting research line is to find architectures for the same task that are more efficient than LocSens. As an example, we have been researching on tagging models that learn a joint embedding space for hashtags and image+location pairs, which at inference time only need to compute a distance between an  image+location  query embedding and pre-computed tags embeddings, being much more efficient. The drawback of such architectures is, however, that the same model cannot be used for tagging and retrieval as LocSens can: A retrieval model with this architecture would have to learn a joint embedding space for hashtags+location pairs and images. 
\vspace{3pt}
\item\textbf{Information modalities balance.} In the paper we propose a location sampling strategy useful to balance the location influence in the image ranking. Experimentation on how this technique can be exploited in other multimodal tasks would be an interesting research line.

\end{itemize}

\end{document}